\pdfoutput=1
\documentclass[sigconf,dvipsnames, svgnames, x11names, natbib=true]{acmart}

\usepackage{tikz}
\usepackage{xcolor}
\usetikzlibrary{arrows}
\usepackage{graphicx}
\usepackage{subfigure}
\usetikzlibrary{shapes}
\usepackage{pgfplots}

\usepackage{graphicx}
\usepackage{amsmath}
\usepackage{amsthm}
\usepackage{booktabs}
\usepackage{algorithm}
\usepackage{algorithmic}
\usepackage{multirow}
\usepackage{amsfonts} %
\def\modelname{LR-Transformer}

\usepackage[misc]{ifsym}
\usepackage{bbding}
\AtBeginDocument{%
  \providecommand\BibTeX{{%
    \normalfont B\kern-0.5em{\scshape i\kern-0.25em b}\kern-0.8em\TeX}}}

\copyrightyear{2021}
\acmYear{2021}
\setcopyright{acmcopyright}
\acmConference[CIKM '21] {Proceedings of the 30th ACM International Conference on Information and Knowledge Management}{November 1--5, 2021}{Virtual Event, Australia.}
\acmBooktitle{Proceedings of the 30th ACM Int'l Conf. on Information and Knowledge Management (CIKM '21), November 1--5, 2021, Virtual Event, Australia}
\acmPrice{15.00}
\acmISBN{978-1-4503-8446-9/21/11}
\acmDOI{10.1145/3459637.3482229}

\begin{document}
\fancyhead{}
\title{An Effective Non-Autoregressive Model for Spoken Language Understanding}


\author{Lizhi Cheng}
\email{clz19960630@sjtu.edu.cn}
\affiliation{%
  \institution{Department of Computer Science and Engineering,\\ Shanghai Jiao Tong University}
  \city{Shanghai}
  \country{PR China}
}

\author{Weijia Jia}
\authornotemark[1]
\email{jiawj@sjtu.edu.cn}
\affiliation{%
  \institution{
    BNU-UIC Institute of Artificial Intelligence and Future Networks Beijing Normal University (BNU Zhuhai),
    Guangdong Key Lab of AI and Multi-Modal Data Processing,
    BNU-HKBU United International College}
  \city{Zhuhai}
  \state{Guangdong}
  \country{PR China}
}

\author{Wenmian Yang}
\authornotemark[1]
\email{dcsyawen@nus.edu.sg}
\affiliation{%
  \institution{Department of Computer Science and Engineering,\\ Shanghai Jiao Tong University}
  \city{Shanghai}
  \country{PR China}
}


\renewcommand{\shortauthors}{}

\begin{abstract}
Spoken Language Understanding (SLU), a core component of the task-oriented dialogue system, expects a shorter inference latency due to the impatience of humans. Non-autoregressive SLU models clearly increase the inference speed but suffer uncoordinated-slot problems caused by the lack of sequential dependency information among each slot chunk. To gap this shortcoming, in this paper, we propose a novel non-autoregressive SLU model named Layered-Refine Transformer, which contains a Slot Label Generation (SLG) task and a Layered Refine Mechanism (LRM). SLG is defined as generating the next slot label with the token sequence and generated slot labels. With SLG, the non-autoregressive model can efficiently obtain dependency information during training and spend no extra time in inference. LRM predicts the preliminary SLU results from Transformer's middle states and utilizes them to guide the final prediction. Experiments on two public datasets indicate that our model significantly improves SLU performance (1.5\% on Overall accuracy) while substantially speed up (more than 10 times) the inference process over the state-of-the-art baseline.
\end{abstract}

\ccsdesc[500]{Computing methodologies~Natural language processing}

\keywords{	
Multi-Task Learning,
Spoken Interfaces,
Task-Oriented Dialogue System}


\maketitle

\section{Introduction}
With the widespread application of intelligent voice assistants, e.g., Apple Siri,task-oriented dialogue systems have received more attention. Working as the spoken interface between users and machines, Spoken Language Understanding (SLU) plays a critical role in the task-oriented dialogue system. A typical SLU task mainly includes two subtasks, i.e., Intent Detection (ID) and Slot Filling (SF) \cite{SLU2011}. Given by an utterance expressed in natural language from the user, ID aims to identify the intent of the user (e.g., GetWeather), and SF aims to fill the slot for each token in the utterance (e.g., location, time). Generally, ID works on sentence-level and is treated as a semantic classification task, while SF is a sequence labeling task focusing on token-level. A simple example of SLU is shown in Figure \ref{fig:example}.

\begin{figure}[h]
\centering
\tikzstyle{background1} = [rectangle, rounded corners=1mm,minimum width = 9.5cm, minimum height = 2.8cm, text centered, draw = black]

\tikzstyle{word} = [rectangle,rounded corners=0.5mm,minimum width = 1.3cm, minimum height = 0.75cm, text centered, thick]
\tikzstyle{slot} = [rectangle,rounded corners=0.5mm,minimum width = 1.3cm, minimum height = 0.75cm, text centered, thick]

\begin{tikzpicture}[node distance = 0cm]

\tikzstyle{every node}=[scale=0.8]
\node(word1)[word,xshift=0cm,yshift=4.5cm]{What};
\node(word1_2)[word,right of=word1,xshift=1cm,yshift=0]{is};
\node(word2)[word,right of=word1_2,xshift=1cm,yshift=0]{the};
\node(word3)[word,right of=word2,xshift=1cm,yshift=0]{weather};
\node(word4)[word,right of=word3,xshift=1cm,yshift=0]{here};
\node(word5)[word,right of=word4,xshift=1cm,yshift=0]{on};
\node(word6)[word,right of=word5,xshift=1cm,yshift=0]{2/7/2021};
\node()[left of=word1,xshift=-1.5cm,yshift=0cm]{\textbf{Utterance}:};

\node(slot1)[slot,below of=word1,xshift=0cm,yshift=-1.1cm]{O};
\node(slot1_2)[slot,right of=slot1,xshift=1cm,yshift=0]{O};
\node(slot2)[slot,right of=slot1_2,xshift=1cm,yshift=0]{O};
\node(slot3)[slot,right of=slot2,xshift=1cm,yshift=0]{O};
\node(slot4)[slot,right of=slot3,xshift=1cm,yshift=0]{B-location};
\node(slot5)[slot,right of=slot4,xshift=1cm,yshift=0]{O};
\node(slot6)[slot,right of=slot5,xshift=1cm,yshift=0]{B-time};
\node(slots)[left of=slot1,xshift=-1.8cm,yshift=0cm ]{\textbf{Slots}:};
\node()[below of=slots,xshift=0.075cm,yshift=-0.7cm ]{\textbf{Intent}:};
\node()[below of=slot1,xshift=0cm,yshift=-0.7cm ]{ GetWeather};

\node()[background1,below of=slot2,xshift=0.1cm,yshift=0.2cm ]{};

\draw[-latex,thin] (word1.south) -- (slot1.north); 
\draw[-latex,thin] (word1_2.south) -- (slot1_2.north);
\draw[-latex,thin] (word2.south) -- (slot2.north); 
\draw[-latex,thin] (word3.south) -- (slot3.north);
\draw[-latex,thin] (word4.south) -- (slot4.north); 
\draw[-latex,thin] (word5.south) -- (slot5.north); 
\draw[-latex,thin] (word6.south) -- (slot6.north); 
\end{tikzpicture}
\caption{An example of SLU.}
\label{fig:example}
\end{figure}

Recent studies find that ID and SF are closely related and positively promote the predicting performance of each other. Specifically, some joint models \cite{JointModel2016,chen2016joint,attention-basedRnn2016} apply a joint loss function to link two tasks, or utilize hidden states of one task in the other \cite{Slot-gated2018,JointCapsule2019,bi-dictional2019}. \citet{qin2019stack}  propagate the token-level intent results to the SF task, achieving the state-of-the-art (SOTA) performance. However, most of the work relies on the RNN-based framework, which models the entire sequence dependency in an autoregressive way, leading to redundant computation and inevitable high latency.

\begin{figure}[h]
\centering
\tikzstyle{background1} = [rectangle, rounded corners=3mm,minimum width = 14cm, minimum height = 5cm, text centered, draw = black, fill = gray!8]
\tikzstyle{memory_list1} = [rectangle,rounded corners=1mm,minimum width = 4cm, minimum height = 0.6cm, text centered, draw = Peru, fill =Peach!15,thick]
\tikzstyle{memory_list2} = [rectangle,rounded corners=1mm,minimum width = 4cm, minimum height = 1.5cm, text centered, draw = Peru, fill =Peach!15,thick]
\tikzstyle{SLU_model} = [rectangle,rounded corners=1mm,minimum width = 1.6cm, minimum height = 0.6cm, text centered, draw = SeaGreen, fill = SeaGreen!15,thick]
\tikzstyle{input} = [rectangle,rounded corners=0mm,minimum width = 4.3cm, minimum height = 0.65cm, text centered, draw = Teal, fill = CornflowerBlue!15,thick]
\tikzstyle{output} = [rectangle,rounded corners=0mm,minimum width = 4.3cm, minimum height = 0.65cm, text centered, draw = Thistle, fill = Thistle!15,thick]

\tikzstyle{word} = [rectangle,rounded corners=0.5mm,minimum width = 1.72cm, minimum height = 0.75cm, text centered, draw = SkyBlue, fill = CornflowerBlue!15,thick]
\tikzstyle{slot} = [rectangle,rounded corners=0.5mm,minimum width = 1.72cm, minimum height = 0.75cm, text centered, draw = Peru, fill = Peach!15,thick]

\tikzstyle{wrong} = [rectangle,rounded corners=0.25mm,minimum width = 4cm, minimum height = 1.15cm, draw =red, dashed]
\tikzstyle{wrong2} = [rectangle,rounded corners=0.25mm,minimum width = 4cm, minimum height = 1.15cm, draw =red, thick]

\begin{tikzpicture}[node distance = 0cm]

\tikzstyle{every node}=[scale=0.7]
\node(CENTER){};
\node(word1)[word,below of=CENTER,xshift=-2.5cm,yshift=1.7cm]{Return};
\node(word2)[word,right of=word1,xshift=2cm,yshift=0]{New};
\node(word3)[word,right of=word2,xshift=2cm,yshift=0]{York};
\node(word4)[word,right of=word3,xshift=2cm,yshift=0]{at};
\node(word5)[word,right of=word4,xshift=2cm,yshift=0]{9};
\node(word6)[word,right of=word5,xshift=2cm,yshift=0]{o'clock};

\node(wslot1)[slot,below of=word1,xshift=0cm,yshift=-1.2cm]{O};
\node(wslot2)[slot,right of=wslot1,xshift=2cm,yshift=0]{B-city};
\node(wslot3)[slot,right of=wslot2,xshift=2cm,yshift=0]{I-time};
\node(wslot4)[slot,right of=wslot3,xshift=2cm,yshift=0]{O};
\node(wslot5)[slot,right of=wslot4,xshift=2cm,yshift=0]{B-city};
\node(wslot6)[slot,right of=wslot5,xshift=2cm,yshift=0]{I-time};
\node()[below of=word4,xshift=-1cm,yshift=1cm]{\textbf{Uncoordinated Slots}:};

\node(word1_2)[word,below of=wslot1,xshift=0cm,yshift=4.5cm]{Return};
\node(word2_2)[word,right of=word1_2,xshift=2cm,yshift=0]{New};
\node(word3_2)[word,right of=word2_2,xshift=2cm,yshift=0]{York};
\node(word4_2)[word,right of=word3_2,xshift=2cm,yshift=0]{at};
\node(word5_2)[word,right of=word4_2,xshift=2cm,yshift=0]{9};
\node(word6_2)[word,right of=word5_2,xshift=2cm,yshift=0]{o'clock};
\node()[below of=word4_2,xshift=-1cm,yshift=1cm]{\textbf{Corret Slots}:};

\node(slot1)[slot,below of=word1_2,xshift=0cm,yshift=-1.2cm]{O};
\node(slot2)[slot,right of=slot1,xshift=2cm,yshift=0]{B-city};
\node(slot3)[slot,right of=slot2,xshift=2cm,yshift=0]{I-city};
\node(slot4)[slot,right of=slot3,xshift=2cm,yshift=0]{O};
\node(slot5)[slot,right of=slot4,xshift=2cm,yshift=0]{B-time};
\node(slot6)[slot,right of=slot5,xshift=2cm,yshift=0]{I-time};


\node(wrong_I)[wrong,below of=wslot2,xshift=1cm]{};
\node(PI1)[below of=wrong_I,yshift=-1cm]{
{B-tag:\color{SeaGreen} \textbf{correct}}
};
\node(PI2)[below of=wrong_I,yshift=-1.5cm]{
{I-tag:\color{red} \textbf{wrong}}
};

\node(wrong_B)[wrong,below of=wslot5,xshift=1cm]{};
\node(PB2)[below of=wrong_B,yshift=-1cm]{
{B-tag:\color{red} \textbf{wrong}}
};
\node(PB2)[below of=wrong_B,yshift=-1.5cm]{
{I-tag:\color{SeaGreen} \textbf{correct}}
};

\draw[-latex,thin] (word1.south) -- (wslot1.north); 
\draw[-latex,thin] (word2.south) -- (wslot2.north); 
\draw[-latex,thin] (word3.south) -- (wslot3.north);
\draw[-latex,thin] (word4.south) -- (wslot4.north); 
\draw[-latex,thin] (word5.south) -- (wslot5.north); 
\draw[-latex,thin] (word6.south) -- (wslot6.north); 

\draw[-latex,thin] (word1_2.south) -- (slot1.north); 
\draw[-latex,thin] (word2_2.south) -- (slot2.north); 
\draw[-latex,thin] (word3_2.south) -- (slot3.north);
\draw[-latex,thin] (word4_2.south) -- (slot4.north); 
\draw[-latex,thin] (word5_2.south) -- (slot5.north); 
\draw[-latex,thin] (word6_2.south) -- (slot6.north); 
\end{tikzpicture}
\caption{An example of uncoordinated slot problem.}
\label{fig:example-uc}
\end{figure}
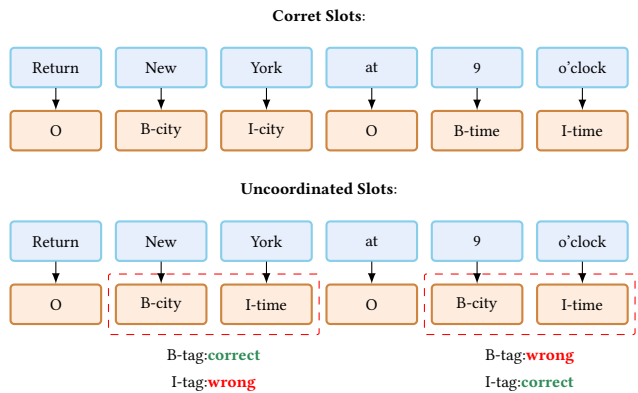

An intuitive idea for accelerating the predicting process is to utilize the self-attention (i.e., Transformer) \cite{vaswani2017attention} instead of the RNN as the basic framework in a non-autoregressive way.
Although using a self-attention based framework does increase the calculation speed, this method will reduce the prediction performance of the SF task \cite{wu2020slotrefine}.
The reason is that SF is a sequence labeling task whose results utilize the “Inside–Outside–Beginning (IOB)” tagging format. Therefore, SF heavily depends on the strongness of the sequential dependency information among each slot chunk (proved by \cite{ren2020studyNAT}). However, the slot labels are predicted independently and simultaneously in self-attention based methods, which reduces the strongness of the sequential dependency and causes the uncoordinated slot problem. Figure \ref{fig:example-uc} shows an example of the uncoordinated slot problem, where \texttt{B-city} should be followed by \texttt{I-city}, but \texttt{I-time} is predicted  by mistake. Therefore, how to maintain the non-autoregressive efficiency when inferring while increase the sequential dependency information to avoid the uncoordinated problem in SLU tasks is a major challenge.

Although \citet{wu2020slotrefine} tries to handle this problem via a two-pass mechanism called SlotRefine, this method still has the following weakness:
First, SlotRefine supposes all the uncoordinated slot problems are caused by wrong `\emph{I-tags}' and only utilize `\emph{B-tags}' to correct the `\emph{I-tags}' in its second pass period. 
However, as shown in Figure \ref{fig:example-uc}, when predicting, an incorrect `\emph{B-tag}' follows by a correct `\emph{I-tag}' also happens. In this case, SlotRefine cannot utilize `\emph{I-tags}' to correct the `\emph{B-tags}'.
Second, the two-pass mechanism needs to characterize text and predict two times in both the training and testing process, which inevitably leads to lower efficiency.
Therefore, how to use the prediction results for interactive feedback to improve the overall performance while avoiding inferring the entire model repeatedly is also a significant challenge.


To solve the above challenges, in this paper, we propose a Layered-Refine Transformer (LR-Transformer) framework for SLU, which containing a Slot Label Generation (SLG) auxiliary task and a Layered Refine Mechanism (LRM) based on the Transformer.
Specifically, SLG is an auxiliary task defined as predicting the next SF label according to the utterance sequence and generated SF labels, which is an autoregressive process like the machine translation. We jointly train the SLG and the original sequence labeling-based SLU tasks by multi-task learning and share their encoder. By SLG, the shared encoder learns more sequential dependency information from the decoder through the cross attention mechanism in the Transformer and further improves sequence labeling-based SLU tasks. Notable, SLG is only employed in the training process and consumes no extra inference time. Thus our model is still a fast non-autoregressive model.
LRM modifies the hidden states between Transformer layers according to the intermediate predicted ID and SF results to help the final prediction. 
More specifically, LRM predicts the preliminary results of ID and SF by the hidden states of the former Transformer layer, merging the result embeddings with the hidden states and inputting them into the next Transformer layer. We have further proved in experiments that LRM only needs to be employed once and does not need to be used between every Transformer layer, so the cost is much smaller than running the entire model twice.


The main contributions of this paper are presented as follows:
\begin{enumerate}
\item We design the SLG as auxiliary multitasking of SLU, which increases the sequential dependency information of the model while consuming no extra inference time.
\item We propose the LRM, which improves the overall performance of SLU tasks by the interaction of predicted results between Transformer layers. Compared with running the entire model twice, the cost of LRM is much smaller.
\item Experimental results on two public datasets show our model is superior to both existing SOTA autoregressive and non-autoregressive models in terms of speed and performance, indicating that our model has great potential for real-world application.
\end{enumerate}

\section{Related Work}
In this section, we introduce some joint learning and multi-task learning approaches utilized in SLU. 

In SLU, intent detection is usually seen as a semantic classification problem to predict the intent label, and slot filling is mainly regarded as a sequence labeling task.
Early studies \cite{yao2014lstm,mesnil2014rnn,peng2015rnn,kurata2016lstm} usually regard ID and SF as two separate tasks and utilize pipeline approaches to manage these two tasks. These methods typically suffer from error propagation due to their independent models and prove less effective than the joint models.

Recently, some work finds that ID and SF are closely related. \citet{Slot-gated2018} propose a slot-gated model that learns the relationship between intent and slot by the gate mechanism. Inspired by \cite{Slot-gated2018}, some bi-directional networks are proposed \cite{bi-dictional2019,liu2019cm}, which dig into the correlation between ID and SF deeper and model the relationship between them more explicitly. In the above work, the interrelated connections between ID and SF are established. Besides, \citet{JointCapsule2019} utilize a hierarchical capsule neural network structure encapsulating the hierarchical relationship among utterance, slot, and intent. \citet{qin2019stack} propose a Stack-Propagation framework that uses result information of ID to guide the SF task but ignores the impact of SF results on the ID task. Although Stack-Propagation reaches the SOTA performance on two SLU datasets, it surfers a long inference latency caused by the heavy and complex framework. \citet{qin2020co} then concern about the impact of SF results in their model but still suffering the long inference latency. Inspired by \cite{yang2019}, \citet{clzICME2021} propose a portable framework RPFSLU, which contains a two-round predicting period. RPFSLU utilizes the semantic information of the first round prediction results to guide the second round prediction by a represent learning process. However, although RPFSLU is portable, it is not a lightweight model due to its two-round prediction process.

The above methods mostly use the autoregressive models (e.g., LSTM and GRU \cite{gru}) that suffer inevitable long inference latency. Inspired by the well performance of Transformer in orther tasks 
\cite{liu2020regularized,li2021ode,li2021learning,}, \citet{wu2020slotrefine} then propose a non-autoregressive joint model named SlotRefine for SLU, which speeds up the predicting process and encounters the uncoordinated-slot problem. SlotRefine tries to handle this problem by a two-pass refine mechanism and get some effect. However, it still suffers uncoordinated slot problems caused by the wrong `I-tag' and not efficient enough during inference due to its two-pass mechanism that needs to run the whole model twice. 

Meanwhile, some work \cite{Multi-domainJoint2016,Multi-TaskNetworks2019} tries to enhance the performance via multi-task learning. 
These multi-task learning methods link the two tasks implicitly via applying a joint loss function.
\citet{memory2019} propose an impressive multi-task learning approach for multi-turn SLU by consolidating context memory with a dialogue logistic inference task called DLI. DLI needs no extra labeled data and only needs to be carried out during training, which inspired us a lot for designing SLG.

Compared with previous work, our LR-Transformer model obtains the sequential dependency information via auxiliary multitasking and utilizes the interaction of the prediction results between ID and SF via the Transformer's middle states. 
Thus, our model has high predicting accuracy and fast inference speed, which indicates that our model has great potential for industrial application.

\section{Method}
In this section, we first introduce the basic model of our LR-Transformer. Then, we descirbe the detail of our Slot Label Generation task. Finally, we introduce the Layered Refine Mechanism. 
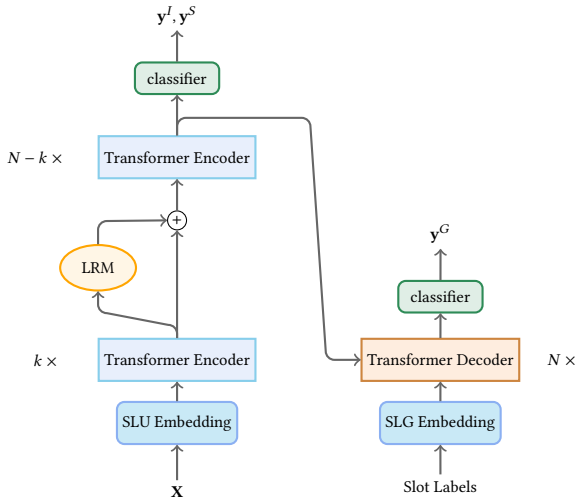
\begin{figure}[h]
\centering
\tikzstyle{background1} = [rectangle, rounded corners=3mm,minimum width = 7cm, minimum height = 3cm, text centered, draw = black, fill = gray!8]
\tikzstyle{background2} = [rectangle, rounded corners=3mm,minimum width = 7cm, minimum height = 8cm, text centered, draw = black, fill = gray!8]

\tikzstyle{Embedding} = [rectangle,rounded corners=1mm,minimum width = 2cm, minimum height = 0.8cm, text centered, draw = CornflowerBlue!75, fill = SkyBlue!45,thick]
\tikzstyle{Encoder} = [rectangle,rounded corners=0mm,minimum width = 2cm, minimum height = 0.8cm, text centered, draw = SkyBlue, fill = CornflowerBlue!15,thick]
\tikzstyle{Decoder} = [rectangle,minimum width = 2cm, minimum height = 0.8cm, text centered, draw = Peru, fill = Peach!15,thick]
\tikzstyle{LRM} = [ellipse,inner sep=2mm, text centered, draw = Orange,fill = Orange!10,thick]
\tikzstyle{classifier} = [rectangle,rounded corners=1mm,minimum width = 1.6cm, minimum height = 0.6cm, text centered, draw = SeaGreen, fill = SeaGreen!15,thick]
\tikzstyle{add} = [circle,minimum size =0.3cm,inner sep=0pt, text centered, draw = black]

\begin{tikzpicture}[node distance = 0cm,scale=7/6]
\tikzstyle{every node}=[scale=0.7]

\node(Embedding_e)[Embedding]{SLU Embedding};
\node(input_e)[below of=Embedding_e,yshift=-1.3cm]{\textbf{X}};
\node(Encoder)[Encoder,below of=Embedding_e,yshift=1.2cm]{Transformer Encoder};
\node()[right of=Encoder,xshift=-2.5cm]{$k\ \times$};
\node(LRM)[LRM,below of=Encoder,xshift=-1.5cm,yshift=1.75cm]{LRM};
\node(add)[add,below of=LRM,xshift=1.5cm,yshift=0.9cm,scale=1.2]{+};
\node(Encoder2)[Encoder,below of=add,yshift=1.2cm]{Transformer Encoder};
\node()[right of=Encoder2,xshift=-2.7cm]{$N-k\ \times$};
\node(liner_e)[classifier,below of=Encoder2,yshift=1.5cm]{classifier};
\node(y_slu)[below of=liner_e,yshift=1.2cm]{$\textbf{y}^{I},\textbf{y}^{S}$};

\node(Embedding_d)[Embedding,below of=Embedding_e,xshift=5cm]{SLG Embedding};
\node(input_d)[below of=Embedding_d,yshift=-1.2cm]{Slot Labels};
\node(Decoder)[Decoder,below of=Embedding_d,yshift=1.2cm]{Transformer Decoder};
\node()[right of=Decoder,xshift=2.3cm]{$N\ \times$};
\node(liner_d)[classifier,below of=Decoder,yshift=1.2cm]{classifier};
\node(y_slg)[below of=liner_d,yshift=1.2cm]{$\textbf{y}^{G}$};

\draw[->,thick,Black!60](input_e.north) -- (Embedding_e.south);
\draw[->,thick,Black!60](Embedding_e.north) -- (Encoder.south);
\draw[->,thick,Black!60](Encoder.north) 
-- ([yshift=0.05cm]Encoder.north) arc(0:90:0.1cm)
-- ([xshift=0.1cm,yshift=-0.25cm]LRM.south) arc(270:180:0.1cm)
-- (LRM.south);
\draw[->,thick,Black!60](LRM.north) -- ([yshift= 0.165cm]LRM.north) arc(180:90:0.1cm) -- (add.west);
\draw[->,thick,Black!60](Encoder.north) -- (add.south);
\draw[->,thick,Black!60](add.north) -- (Encoder2.south);
\draw[->,thick,Black!60](Encoder2.north) -- (liner_e.south);
\draw[->,thick,Black!60](liner_e.north) -- (y_slu.south);

\draw[->,thick,Black!60](Encoder2.north) 
--([yshift=0.1cm]Encoder2.north) arc(180:90:0.1cm)
--([xshift=1.42cm,yshift=0.2cm]Encoder2.north) arc (90:0:0.1cm)
--([xshift=-0.4cm,yshift=0.1cm]Decoder.west)arc (180:270:0.1cm)
--(Decoder.west);

\draw[->,thick,Black!60](input_d.north) -- (Embedding_d.south);
\draw[->,thick,Black!60](Embedding_d.north) -- (Decoder.south);
\draw[->,thick,Black!60](Decoder.north) -- (liner_d.south);
\draw[->,thick,Black!60](liner_d.north) -- (y_slg.south);

\end{tikzpicture}
\caption{General framework.}
\label{fig:model}
\end{figure}

\subsection{Problem Formulation and Basic Model}
\label{sec:basic-model}
In this section, we introduce the problem formulation and describe our basic model in detail.

The input of SLU tasks is an utterance composed by a token sequence $\textbf{X}=\{x_1,...,x_n\}$, where $n$ donates the sequence length.
Given $\textbf{X}$ as input, our tasks are composed of Intent Detection (ID) and Slot Filling (SF).
Specifically, ID is a semantic classification task to predict the intent label for the whole utterance, while SF is a sequence labeling task to give each token in the sequence a slot label. In our model, the intent label and all slot labels are predicted simultaneously.

Following previous non-autoregressive models, we employ the multi-head Transformer encoder \cite{vaswani2017attention} as our basic model.
\citet{vaswani2017attention} describe the Transformer framework in great detail, so we do not need to introduce it again. The only difference is that we utilize self-attention with relative position representations \cite{shaw2018position_relative_attention} to address the sequential information.

Akin to the operation in BERT \cite{BERT}, we first insert a special token `\emph{CLS}' in the beginning of the token sequence, which is utilized to predict the label of the intent. Given the new sequence $\textbf{X}=\{x_{cls},x_1,...,x_n\}$ as input, the Transformer encoder returns the hidden states sequence $\textbf{H}=\{\textbf{h}_{cls},\textbf{h}_1,...,\textbf{h}_n \in \mathbb{R}^{ d_{model} }\}$ as output, where $d_{model}$ is the input and output dimension of the Transformer layer. Then, the prediction of ID and SF are calculated as:
\begin{equation}
\begin{aligned}
    \textbf{y}^I &= \mathrm{softmax} ( \textbf{W}^I \cdot \textbf{h}_{cls} + \textbf{b}^I )\\
    \textbf{y}^S_j &= \mathrm{softmax} ( \textbf{W}^S \cdot (\textbf{h}_j \oplus \textbf{h}_{cls}) + \textbf{b}^S )
\end{aligned}
\label{eq:getres}
\end{equation}
where $\textbf{y}^I \in \mathbb{R}^{d_{i}} $ and $\textbf{y}^S = \{ \textbf{y}^S_1,..., \textbf{y}^S_n \in \mathbb{R}^{d_{s}} \}$ donate the results of ID and SF, $d_{i}$ and $d_s$ are the the categories of the intent label and slot labels,
$\textbf{W}^I \in \mathbb{R}^{ d_{i} \times d_{model} }$ and $\textbf{W}^S \in \mathbb{R}^{ d_{s} \times 2d_{model}}$ are fully connected matrices, 
$\textbf{b}^I \in \mathbb{R}^{d_{i}}$ and $\textbf{b}^S \in \mathbb{R}^{d_{s}}$ are bias vectors, and $\oplus$ donates the concatenation operation.

The objective of our basic model can be formulated as:
\begin{equation}
    p\left(\textbf{y}^{I}, \textbf{y}^{S} \mid \textbf{X}\right)=p\left(\textbf{y}^{I} \mid \textbf{X}\right) \cdot \prod_{j}^{n} p\left(\textbf{y}_{j}^{S} \mid \textbf{X}, \textbf{y}^{I}\right)
\end{equation}

The joint loss function of SLU is defined as:
\begin{equation}
\begin{aligned}
\mathcal{L}_{SLU} &=-\log P(\textbf{y}^{I} \mid x_{1}, \ldots, x_{n}) \\
&-\sum^{n}_{j=1} \log P(\textbf{y}_{j}^{S} \mid x_{1}, \ldots, x_{n})
\end{aligned}
\end{equation}

\subsection{Slot Label Generation}
\label{sec:SLG}
In this section, we introduce the Slot Label Generation (SLG) task in detail.

As we aforementioned, the SF task heavily depends on the strongness of the sequential dependency information among each slot chunk. Although we have tried to obtain dependency information via self-attention with relative position representations \cite{shaw2018position_relative_attention}, it is not enough for the sequence labeling task like SF. 

To obtain more sequential dependency information for the model, we design the SLG as auxiliary multitasking, jointly training the original sequence labeling-based SLU tasks and sharing their encoder.

Specifically, given the token sequence $\textbf{X}=\{x_{cls},x_1,...,x_n\}$ and previous predicted slot labels $\{ \textbf{y}^G_{1},..., \textbf{y}^G_{j-1} \}$, SLG generates the next slot label  $\textbf{y}^G_j$ by an autoregressive way. The objective of our SLG can be formulated as:
\begin{equation}
\begin{aligned}
    p( \textbf{y}^G \mid \textbf{X})= \prod_{j=1}^{n} p(\textbf{y}_{j}^{G} \mid \textbf{X}, \textbf{y}_{1}^{G},...,\textbf{y}_{j-1}^{G}  )
\end{aligned}
\end{equation}
where $\textbf{y}^G = \{ \textbf{y}^G_1,..., \textbf{y}^G_n \in \mathbb{R}^{d_{s}} \}$.

In practice, we extend our basic model with a Transformer decoder \cite{vaswani2017attention} to construct the model architecture of SLG. The complete framework of SLG works as a sequence to sequence (seq2seq) model. Since the Transformer-based seq2seq model has been widely used, we do not describe it in more detail.

Moreover, to enhance the prediction consistency between the SLU task and the SLG task, we further design a consistency loss function based on the cross entropy.
The final loss function of SLG is defined as:

\begin{equation}
\begin{aligned}
\mathcal{L}_{SLG} 
&= (1-\alpha) \cdot -\log P( \textbf{y}^{G} \mid\textbf{X} ) +\alpha \cdot H(  \textbf{y}^{G} ,\textbf{y}^{S} )\\
&= (\alpha-1) \cdot \log P( \textbf{y}^{G} \mid\textbf{X} ) +\alpha \cdot H(  \textbf{y}^{G} ,\textbf{y}^{S} )
\label{eq_slgloss}
\end{aligned}
\end{equation}
where $H$ is the cross-entropy function, $\alpha$ is a hyper-parameter and $\textbf{y}^{S}$ is the predicted labels of SF obtained by Eq.\ref{eq:getres}.  

By SLG, the shared encoder can learn more sequential dependency information from the decoder through the cross attention mechanism in Transformer and further improve sequence labeling-based SLU tasks. 

The loss function of overall multi-task learning is defined as:
\begin{equation}
    \mathcal{L}= \mathcal{L}_{SLU}+\lambda \mathcal{L}_{SLG}
\label{eq:jloss}
\end{equation}
where $\lambda$ is a hyper-parameter.

Notably, the SLG task is only carried out during the training process and
costs no extra time for inference. Therefore, our SLU model is still a non-autoregressive model, which is very efficient.

\subsection{Layered Refine Mechanism}
\label{sec:LRM}
In this section, we introduce the Layered Refine Mechanism (LRM) for the Transformer in detail.

\begin{figure}[h]
\centering
\tikzstyle{background1} = [rectangle, rounded corners=3mm,minimum width = 7cm, minimum height = 3cm, text centered, draw = black, fill = gray!8]
\tikzstyle{background2} = [rectangle, rounded corners=3mm,minimum width = 7cm, minimum height = 8cm, text centered, draw = black, fill = gray!8]

\tikzstyle{blank} = [rectangle, rounded corners=1mm,minimum width = 5cm, minimum height = 0.8cm, draw = orange,  dashed]

\tikzstyle{Encoder} = [rectangle,rounded corners=2.5mm,minimum width = 8cm, minimum height = 0.8cm, text centered, draw = SkyBlue, fill = CornflowerBlue!15,thick]
\tikzstyle{Encoder_cell} = [rectangle,rounded corners= 0.5mm,minimum width = 1cm, minimum height = 0.6cm, text centered, draw = SkyBlue, fill = CornflowerBlue!15, thick]
\tikzstyle{Encoder_cell_b} = [rectangle,rounded corners= 0.5mm,minimum width = 1cm, minimum height = 0.8cm, text centered, thick]

\tikzstyle{Embedding} = [rectangle,rounded corners=1mm,minimum width = 7.2cm, minimum height = 0.8cm, text centered, draw = Orange, fill = Orange!10,thick]
\tikzstyle{Embedding_cell} = [rectangle,rounded corners=1mm,minimum width = 0.8cm, minimum height = 0.6cm, text centered, draw = Orange, fill = Orange!10,thick]
\tikzstyle{Embedding_cell_b} = [rectangle,rounded corners=1mm,minimum width = 1cm, minimum height = 0.8cm, text centered,thick]

\tikzstyle{Decoder} = [rectangle,minimum width = 2cm, minimum height = 0.8cm, text centered, draw = Peru, fill = Peach!15,thick]
\tikzstyle{LRM} = [ellipse,inner sep=2mm, text centered, draw = Orange,fill = Orange!10,thick]

\tikzstyle{classifier} = [rectangle,rounded corners=1mm,minimum width = 7.2cm, minimum height = 0.8cm, text centered, draw = SeaGreen, fill = SeaGreen!15,thick]
\tikzstyle{classifier_cell} = [rectangle,rounded corners=1mm,minimum width = 0.8cm, minimum height = 0.6cm, text centered, draw = SeaGreen, fill = SeaGreen!15, thick]
\tikzstyle{classifier_cell_b} = [rectangle,rounded corners=1mm,minimum width = 1cm, minimum height = 0.8cm, text centered, thick]

\tikzstyle{add} = [circle,minimum size =0.3cm,inner sep=0pt, text centered, draw = black!40]
\tikzstyle{add_b} = [circle,minimum size =0.3cm,inner sep=0pt, text centered]
\tikzstyle{Function} = [circle,minimum size =0.6cm,inner sep=0pt, text centered, draw = Orange,fill = Orange!10,thick]
\tikzstyle{blank1} = [rectangle,minimum width = 1cm, minimum height = 0.6cm, text centered, draw = white,draw opacity=0, fill = white, fill opacity=0,thick]

\begin{tikzpicture}[node distance = 0cm]
\tikzstyle{every node}=[scale=0.7]

\node(Encoder)[Encoder]{Encoder Layer $k$};
\node(e_cls)[Encoder_cell_b,below of = Encoder,xshift=-3cm,yshift=0cm]{};
\node(e_1)[Encoder_cell_b,right of = e_cls,xshift=2cm,yshift=0cm]{}; 
\node(e_2)[Encoder_cell_b,right of = e_1,xshift=2cm,yshift=0cm]{}; 
\node(e_3)[Encoder_cell_b,right of = e_2,xshift=2cm,yshift=0cm]{};

\node(h_cls)[Encoder_cell, below of = e_cls,xshift=0cm,yshift=1.2cm]{$\textbf{h}^k_{cls}$};
\node(h_1)[Encoder_cell, right of = h_cls,xshift=2cm,yshift=0cm]{$\textbf{h}^k_1$}; 
\node(h_2)[Encoder_cell_b, right of = h_1,xshift=2cm,yshift=0cm]{...}; 
\node(h_3)[Encoder_cell, right of = h_2,xshift=2cm,yshift=0cm]{$\textbf{h}^k_n$};

\node(liner)[classifier,below of = h_cls,xshift=3cm,yshift=1.5cm]{Classifier};
\node(c0)[classifier_cell_b,below of = liner, xshift=-3cm]{};
\node(c1)[classifier_cell_b,below of = c0, xshift=2cm]{};
\node(c2)[classifier_cell_b,below of = c1, xshift=2cm]{};
\node(c3)[classifier_cell_b,below of = c2, xshift=2cm]{};

\node(y0)[classifier_cell,below of = c0, yshift=1.2cm]{$\tilde{\textbf{y}}^{I}$};
\node(y1)[classifier_cell,below of = y0, xshift=2cm]{$\tilde{\textbf{y}}^S_1$};
\node(y2)[classifier_cell_b,below of = y1, xshift=2cm]{...};
\node(y3)[classifier_cell,below of = y2, xshift=2cm]{$\tilde{\textbf{y}}^S_n$};

\node(Embedding)[Embedding,below of = y0,xshift=3cm,yshift=1.5cm]{Result Embedding};
\node(eb0)[Embedding_cell_b,below of = Embedding, xshift=-3cm]{};
\node(eb1)[Embedding_cell_b,below of = eb0, xshift=2cm]{};
\node(eb2)[Embedding_cell_b,below of = eb1, xshift=2cm]{};
\node(eb3)[Embedding_cell_b,below of = eb2, xshift=2cm]{};

\node(re0)[Embedding_cell,below of = eb0, yshift=1.2cm] {$\textbf{e}^{I},\textbf{e}_0^{S}$};
\node(re1)[Embedding_cell,below of = re0, xshift=2cm]{$\textbf{e}^S_1$};
\node(re2)[Embedding_cell_b,below of = re1, xshift=2cm]{...};
\node(re3)[Embedding_cell,below of = re2, xshift=2cm]{$\textbf{e}^S_n$};

\node(re123)[blank,below of = re2]{};


\node(add_I)[add,below of=re0,xshift=0cm,yshift=1.5cm,scale=1.2]{+};
\node(add1)[add_b,right of=add_I,xshift=2cm,yshift=0cm,scale=1.2]{...};
\node(add2)[right of=add1,xshift=2cm,yshift=0cm]{...};
\node(add3)[add,right of=add2,xshift=2cm,yshift=0cm,scale=1.2]{+};

\node(h_cls2)[Encoder_cell, below of = add_I,xshift=0cm,yshift=1.2cm]{${\textbf{h}'}_{cls}$};
\node(h_12)[Encoder_cell, right of = h_cls2,xshift=2cm,yshift=0cm]{$\textbf{h}'_1$}; 
\node(h_22)[Encoder_cell_b, right of = h_12,xshift=2cm,yshift=0cm]{...}; 
\node(h_32)[Encoder_cell, right of = h_22,xshift=2cm,yshift=0cm]{$\textbf{h}'_n$};

\node(Encoder2)[Encoder,below of=h_cls2,xshift=3cm,yshift=1.2cm]{Encoder Layer $k+1$};
\node(e_cls2)[Encoder_cell_b,below of = Encoder2, xshift=-3cm, yshift=0cm]{};
\node(e_12)[Encoder_cell_b,right of = e_cls2, xshift=2cm, yshift=0cm]{}; 
\node(e_22)[Encoder_cell_b,right of = e_12,xshift=2cm,yshift=0cm]{}; 
\node(e_32)[Encoder_cell_b,right of = e_22, xshift=2cm, yshift=0cm]{};

\node(p0)[below of = h_cls, xshift=-1cm]{};
\node(p0_add)[below of = add_I, xshift=-1cm]{};
\draw[->, thin,SkyBlue](h_cls.west) -- (p0.east) -- (p0_add.east) --(add_I.west);


\node(p3)[below of = h_3, xshift=1cm]{};
\node(p3_add)[below of = add3, xshift=1cm]{};
\draw[->, thin,SkyBlue](h_3.east) -- (p3.west) -- (p3_add.west) --(add3.east);

\draw[-latex, thick,SkyBlue](e_cls.north)  -- (h_cls.south);
\draw[-latex, thick,SkyBlue](e_1.north)  -- (h_1.south);
\draw[-latex, thick,SkyBlue](e_3.north)  -- (h_3.south);

\draw[->, thick,SkyBlue](h_cls.north)  -- (c0.south);
\draw[->, thick,SkyBlue](h_1.north)  -- (c1.south);
\draw[->, thick,SkyBlue](h_3.north)  -- (c3.south);

\draw[->, dotted, thick, SkyBlue](h_cls.north)  .. controls ([xshift=0.2cm]h_cls.north) and ([xshift=-0.3cm,yshift=-0.5cm]c1.south) .. ([xshift=-0.1cm,yshift=-0.1cm]c1.south);

\draw[->, dotted, thick, SkyBlue](h_cls.north)  .. controls ([xshift=0.2cm]h_cls.north) and ([xshift=-0.8cm,yshift=-0.5cm]c2.south) .. ([xshift=-0.1cm,yshift=-0.1cm]c2.south);

\draw[->, dotted, thick, SkyBlue](h_cls.north)  .. controls ([xshift=0.2cm]h_cls.north) and ([xshift=-1cm,yshift=-0.5cm]c3.south) .. ([xshift=-0.1cm,yshift=-0.1cm]c3.south);

\draw[->,thick,SeaGreen](c0.north)  -- (y0.south);
\draw[->,thick,SeaGreen](c1.north)  -- (y1.south);
\draw[->,thick,SeaGreen](c3.north)  -- (y3.south);

\draw[->,thick,SeaGreen](y0.north)  -- (eb0.south);
\draw[->,thick,SeaGreen](y1.north)  -- (eb1.south);
\draw[->,thick,SeaGreen](y3.north)  -- (eb3.south);

\draw[->,thick,orange](eb0.north)  -- (re0.south);
\draw[->,thick,orange](eb1.north)  -- (re1.south);
\draw[->,thick,orange](eb3.north)  -- (re3.south);

\draw[->,thick,orange](re0.north)  -- (add_I.south);
\draw[->,thick,orange](re1.north)  -- (add1.south);
\draw[->,thick,orange](re3.north)  -- (add3.south);

\draw[->,thin,orange](re123.west)  -- (re0.east);

\draw[-latex,thick,Black!40](add_I.north) -- (h_cls2.south);
\draw[-latex,thick,Black!40](add1.north) -- (h_12.south);
\draw[-latex,thick,Black!40](add3.north) -- (h_32.south);

\draw[-latex,thick,SkyBlue](h_cls2.north) -- (e_cls2.south);
\draw[-latex,thick,SkyBlue](h_12.north) -- (e_12.south);
\draw[-latex,thick,SkyBlue](h_32.north) -- (e_32.south);

\end{tikzpicture}
\caption{LRM architecture.}
\label{fig:model}
\end{figure}
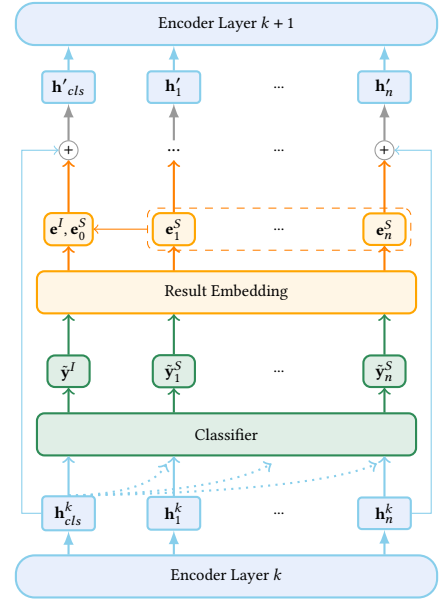


Previous work \cite{clzICME2021} has widely proved that ID and SF are closely related. Taking advantage of the correlation between these two tasks, especially utilizing one task's results in the other, can effectively enhance the overall performance.
However, since the non-autoregressive approach predicts the results of ID and SF simultaneously, we can not directly employ these results in a one-pass prediction process like Stack-Propagation \cite{qin2019stack} do.
Although we can utilize a two-pass mechanism to generate the first-pass results and guide the second-pass prediction via them, it is a trade-off between autoregression and non-autoregression, which costs much time.

Considering Transformer contains a multi-layer architecture, and each encoder layer of the Transformer has the same structure. We design the LRM which works between two Transformer layers and utilizes middle states of the Transformer to guide the final prediction.

Specifically, we first predict a preliminary results of ID $\tilde{\textbf{y}}^I$ and SF $\tilde{\textbf{y}}^S=\{ \tilde{\textbf{y}}^S_1,...,\tilde{\textbf{y}}^S_n \}$ by Eq.\ref{eq:getres} according to hidden states $\textbf{H}^k=\{\textbf{h}^k_{cls},\textbf{h}^k_1,...,\textbf{h}^k_n \}$ from the $k$-th Transformer layer. 
Then, we embed $\tilde{\textbf{y}}^I$ into $\textbf{e}^{I} \in \mathbb{R}^{d_e}$ and $\tilde{\textbf{y}}^S=\{ \tilde{\textbf{y}}^S_1,...,\tilde{\textbf{y}}^S_n \}$ into $\textbf{e}^{S}=\{ \textbf{e}^S_1,...,\textbf{e}^S_n \in \mathbb{R}^{d_e}  \}$ by the embedding layer, respectively, where $d_e$ is the embedding size and we set $d_e=d_{model}$ in this paper.

Since the SF returns a sequence of result embedding vectors, we further calculate the weighted  average of those embedding vectors via an attention mechanism  to obtain an utterance-level result embedding vector $\textbf{e}^S_0 \in \mathbb{R}^{d_e}$ by
\begin{equation}
    \textbf{e}^S_0=\sum_{j=1}^{n} \alpha_{j} \cdot \textbf{e}^S_j
\end{equation}
where $\alpha_j$ is the weight of $\textbf{e}^S_j$ obtained by
\begin{equation}
\alpha_{j}=\frac{\operatorname{exp}(\textbf{e}^S_{j})}{\sum_{k=1}^{n} \operatorname{exp}(\textbf{e}^S_{k})}
\end{equation}

By the above calculation, we obtain result embedding vectors $\textbf{e}^{I}$, $\textbf{e}^{S}$ and $\textbf{e}^{S}_0$, which contain semantic information from the preliminary results of ID and SF. Subsequently, we merge these vectors with the former output and obtain a new hidden states sequence  $\textbf{H}'=\{\textbf{h}'_{cls},\textbf{h}'_1,...,\textbf{h}'_n \}$ by
\begin{align}
    \textbf{h}_{cls}'&= \textbf{h}^k_{cls} + \textbf{e}^{I} + \textbf{e}^{S}_0\\
    \textbf{h}_j'&= \textbf{h}^k_j+\textbf{e}^{S}_j
\end{align}
We use $\textbf{H}'$ as input of the $k+1$-th Transformer layer.

LRM can incorporate the bidirectional semantic information from one task to the other by propagating the combination of former output and preliminary results so that  ID and SF become more accurate.
The complete Markov chain process can be simplified as follow:
\begin{equation}
\begin{aligned}
    p(&\textbf{y}^{I}, \textbf{y}^{S}  \mid \textbf{X}) =p(\tilde{\textbf{y}}^{I} \mid \textbf{X}) \cdot  p(\tilde{\textbf{y}}^{S} \mid \textbf{X}, \tilde{\textbf{y}}^{I})\\
    &\cdot  p(\textbf{y}^{I} \mid \textbf{X}, \tilde{\textbf{y}}^{I}, \tilde{\textbf{y}}^{S} ) \cdot  p(\textbf{y}^{S} \mid \textbf{X}, \tilde{\textbf{y}}^{I}, \tilde{\textbf{y}}^{S}, \textbf{y}^{I} )
\end{aligned}
\end{equation}

LRM is a portable plugin for the Transformer, which can be utilized between two Transformer encoder layer intervals. Actually, using LRM once in the whole Transformer structure is enough because it already considers the interaction between ID and SF. On the contrary, overusing LRM  will make the Transformer layers lose their own feature and negatively affect the final performance because we use the SLU classifier in LRM. Also, LRM is not suitable for the autoregressive SLG task because it uses all SLU prediction labels simultaneously.

Besides, the network structure of LRM is mainly composed of a fully connected layer as the SLU classifier and an embedding layer for result embedding. Therefore, our LRM is very lightweight and consumes very little inference time.

\section{Experiment}

In this section, we demonstrate the effectiveness of \modelname{}. 
We first introduce datasets, the necessary hyper-parameters, and the baselines used in our experiments.
Then, we compare the performance of our framework with baselines and analyze the experiment results. 
Subsequently, we analyze the error caused by the uncoordinated problem and carry out an ablation study to verify the effeteness of SLG and LRM.
Finally, we combine our model with the pre-trained model and analyze the effect.

\subsection{Experimental Settings and Baselines}
\begin{table}[h]
\centering
\begin{tabular}{lrrr}
\toprule
Dataset                 & ATIS      & SNIPS     \\
\midrule
Vocabulary Size         & 722       & 11241     \\
Avg. tokens per utterance    & 11.28     & 9.05      \\
Intent categories             & 21        & 7         \\
Slot categories              & 120       & 72        \\
Training set size            & 4478      & 13084     \\
Validation set size          & 500       & 700       \\
Test set size                 & 893       & 700       \\
\bottomrule
\end{tabular}
\caption{Dataset statistics.}
\label{dataset}
\end{table}

\paragraph{Dataset:}
To evaluate the efficiency of our proposed model, we conduct experiments on two public datasets, i.e., ATIS (Airline Travel Information Systems \cite{atis}) and SNIPS (collected by Snips personal voice assistant \cite{snips}). Compared
with ATIS, the SNIPS dataset is more complex due to its large vocabulary size, cross-domain intents, and more out-of-vocabulary words.
The statistics of ATIS and SNIPS are shown in Table \ref{dataset}.

\paragraph{Evaluation Metrics:}
Following previous work, we evaluate the SLU performance of ID by accuracy and the performance of SF by the F1 score.
Besides, we utilize overall accuracy to indicate the proportion of utterance in the corpus whose slots and intent are both correctly predicted. Usually, a higher intent accuracy and F1 score also lead to higher overall accuracy. However, this does not always happen, e.g., when the prediction contains more mistakes, but most mistakes are from the same utterances.

\paragraph{Set up:}
Following previous work, we use Adam \cite{adam} to optimize the parameters in our model and adopted the suggested learning rate of 0.001. The batch size is set to 32 according to the size of training data.

When tuning hyper-parameters, we repeat the model 5 times and select the parameters with the best average performance on the validation set as the optimal.

We first select the hyper-parameters used in our basic model. To select Transformer input and output size $d_{model}$, and the size of inner-layer in the feed-forward network of Transformer (we call it $d_{ff}$ in the following part), we leverage the grid search. Specifically, we determine $d_{model}$ in the range of $\{128,\ 256,\ 512 \}$ and $d_{ff}$ in the range of $\{ 128,\ 256,\ 512,\ 768\}$. We finally choose $d_{model}$ as 128 and $d_{ff}$ as 512 as the optimal. For other hyper-parameters of Transformer, following \cite{vaswani2017attention}, we set both encoder and decoder layers as 6, the number of attention heads as 8, and the dropout ratio as 0.3. We utilize LRM once between the second and the third Transformer layer.

Then, we choose the hyper-parameters $\alpha$ used in Eq.\ref{eq_slgloss} and $\lambda$ used in Eq.\ref{eq:jloss}. We first fix $\lambda$ as 1 and select $\alpha$ in the range of (0,0.5] with the step 0.05. Subsequently, with the selected $\alpha$, we select $\lambda$ in range of (0,1] with the step 0.25. We finally get the optimal when $\alpha$  is 0.35 and $\lambda$ is 0.75. We will introduce the influence of $\lambda$ and $\alpha$ in our ablation study.

\begin{table*}[t]
\centering
\begin{tabular}{l|ccc|ccc}
\hline
\multirow{2}{*}{\textbf{Model}}  & \multicolumn{3}{c|}{\textbf{ATIS}}    & \multicolumn{3}{c}{\textbf{SNIPS}} \\ 
\cline{2-7} & \multicolumn{1}{c}{Intent} & \multicolumn{1}{c}{Slot} & Overall & \multicolumn{1}{c}{Intent} & \multicolumn{1}{c}{Slot} & Overall \\ 
\hline
\multicolumn{7}{c}{\textbf{Autoregressive Models}} \\ 
\hline
Joint Seq \cite{Multi-domainJoint2016}          & 92.6       & 94.2          & 80.7        & 96.9       & 87.3          & 73.2   \\
Attention-BiRNN \cite{attention-basedRnn2016}   & 91.1       & 94.2          & 78.9        & 96.7       & 87.8          & 74.1   \\
Slot-Gated \cite{Slot-gated2018}    & 93.6       & 94.8          & 82.2        & 97.0       & 88.8          & 75.5   \\
SF-ID \cite{bi-dictional2019}       & 97.8       & 95.8          & 86.8        & 97.4       & 92.2          & 80.6   \\
Stack-Propagation \cite{qin2019stack}           & 96.9       & 95.9          & 86.5        & 98.0       & 94.2          & 86.9   \\

\hline
\multicolumn{7}{c}{\textbf{Non-autoregressive Models}} \\ 
\hline
SlotRefine \cite{wu2020slotrefine}  & 97.1       & 96.0          & 86.9        & 97.4       & 93.5          & 84.4   \\ 
Basic model  & 96.8      & 95.2      & 85.6      & 96.1      & 92.8      & 82.1 \\
\modelname{} & \ \textbf{98.2}$\uparrow$ & \ \textbf{96.1}$\uparrow$ & \ \textbf{87.2}$\uparrow$ & \ \textbf{98.4}$\uparrow$ & \ \textbf{94.8}$\uparrow$ & \ \textbf{88.4}$\uparrow$ \\
\hline
\end{tabular}
\caption{SLU performance on ATIS and SNIPS datasets.
The numbers with $\uparrow$ indicate that the improvement of our model over all baselines is statistically significant with $p < 0.05$ under t-test.}
\label{res0}
\end{table*}

\paragraph{Baselines:}
We compare our model with the existing baselines, including:
\begin{itemize}
\item Joint Seq \cite{Multi-domainJoint2016}: A GRU \cite{gru} based model with a multi-task modeling approach.
\item Attention-BiRNN \cite{attention-basedRnn2016}: A LSTM \cite{LSTM} based encoder-decoder model with an intent attention mechanism.
\item Slot-gated \cite{Slot-gated2018}: A LSTM based joint model together with a slot-gated mechanism as a special gate function.
\item SF-ID \cite{bi-dictional2019}: A LSTM based joint model with cross-impact calculating between two tasks.
\item Stack-Propagation \cite{qin2019stack}: A LSTM based joint model with stack-propagation framework and token-level ID. This model has already guided SF by ID results, which is the state-of-the-art of the joint model.
\item Basic model: The encoder of Transformer framework with relative position representations \citet{vaswani2017attention}.  
\item SlotRefine \cite{wu2020slotrefine}: A Transformer based non-autoregressive model with a two-pass refine mechanism. 
\end{itemize}

For Joint Seq, Attention BiRNN, Slot-gated, SF-ID, and Stack-Propagation, we adopt the reported results from \cite{qin2019stack}.
For SlotRefine \cite{wu2020slotrefine}), since the benchmark in their original paper is calculated nonstandardly according to their open-source code, we re-implemented the model (all hyper-parameters strictly identical as \cite{wu2020slotrefine}) and obtained the results. Note that, for SlotRefine, basic model, and LR-Transformer, we repeat the experiment  5 times and report the average as the final results.

\subsection{Result and Analysis}
In this section, we show the results of our experiments and do some analysis.

\paragraph{SLU Performance:}
The experiment results of the proposed models on ATIS and SNIPS datasets are shown in Table \ref{res0}.
The results show that our model significantly outperforms all the baselines and achieves the best performance in all three metrics.
Compared with the prior non-autoregressive model SlotRefine, our model enhances the performance by 1.1\%(ID), 0.1\%(SF), and 0.3\%(Overall) on ATIS and 1.0\%(ID), 1.3\%(SF), and 4.0\%(Overall) on SNIPS. Compared with the SOTA baseline Stack-Propagation, \modelname{} also achieve improvement by 1.3\%(ID), 0.2\%(SF), and 0.7\%(Overall) on ATIS and 0.4\%(ID), 0.6\%(SF), and 1.5\%(Overall) on SNIPS.
This indicates the effectiveness of our \modelname{}.

Notably, without SLG and LRM, our basic model performs worse than both SlotRefine and Stack-Propagation, but \modelname{} outperforms both of them with SLG and LRM. We attribute this enhancement to the fact that our SLG task effectively obtains the sequential dependency information, and LRM directly takes the explicit result information into consideration, which grasps the relationship between the intent and slots. We will conduct experiments for the ablation study in section 4.4 to further verify this idea.

\paragraph{Speed Up:}
The inference time of SLU models is shown in Table \ref{res_speed}. All the models in this experiment are conducted with a single TITAN Xp GPU. 
From the table, we can obviously find that our model achieves significant speedup ($\times$10.53 on ATIS; $\times$10.24 on SNIPS) against the autoregressive SOTA model Stack-Propagation because all slot labels are calculated simultaneously in our non-autoregressive method.

\begin{table}[h]
\centering
\resizebox{\columnwidth}{!}{
\begin{tabular}{l|cc|cc}
\hline
\multirow{2}{*}{Model} & \multicolumn{2}{c|}{\textbf{ATIS}}          & \multicolumn{2}{c}{\textbf{SNIPS}}         \\ \cline{2-5} 
                       & \multicolumn{1}{c}{Latency } & Speedup & \multicolumn{1}{c}{Latency } & Speedup \\ 
\hline
Stack-Propagation      & 138.12ms   & 1.00$\times$  & 143.44ms  & 1.00$\times$   \\
SlotRefine             & 32.04ms    & 4.31$\times$  & 41.12ms   & 3.48$\times$   \\ 
\hline
\modelname{} w/o LRM    & 12.93ms   & 10.68$\times$  & 13.82ms  & 10.38$\times$ \\
\modelname{}    & 13.11ms   & 10.53$\times$  & 14.01ms  & 10.24$\times$  \\
\hline
\end{tabular}
}
\caption{Latency of SLU models. ``Latency'' is the average inference time without minibatching. “Speedup” is compared against the existing SOTA model Stack-Propagation \cite{qin2019stack}.}
\label{res_speed}
\end{table}

More importantly, compared with the existing non-autoregressive model SlotRefine, our model reduces nearly 60\% inference latency on SNIPS and more than 65\% on ATIS.
The reason for this phenomenon is that SlotRefine needs to run the whole model twice, including the embedding layer, Transformer, and the classifier. As a comparison, our model is lightweight and obtains the final SLU results with a one-period prediction. Thus, although our model contains more transformer layers, it is still significantly faster than SlotRefine.
From Table \ref{res_speed}, we can also find that utilizing LRM
consumes only 3\% extra inference time compared to without LRM. 
This indicates that LRM is a lightweight approach, which generates negligible time cost. 

\subsection{Error Analysis}
In this section, we will analysis the error caused by uncoordinated slots.

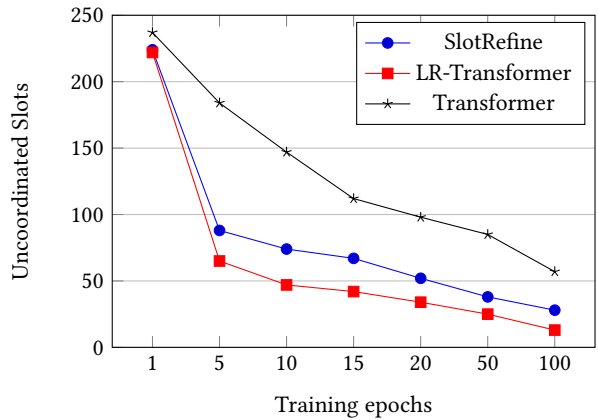
\begin{figure}[h]
\centering
\begin{tikzpicture}[scale = 1]
\begin{axis}[
height=6cm,
width=8cm,
xlabel=Training epochs,
ylabel=Uncoordinated Slots,
ymin=0,
ymax=250,
ytick pos=left,
symbolic x coords={1,5,10,15,20,50,100},
ytick={0,50,100,150,200,250},
ymajorgrids=true,
x tick label style={/pgf/number format/1000 sep={}},
]
\addplot coordinates {
(1, 224)
(5,	88)
(10, 74)
(15,67)
(20,52)
(50,38)
(100,28)
};
\addlegendentry{SlotRefine}


\addplot [color=red,mark=square*]
coordinates {
(1, 222)
(5,	 65)
(10, 47)
(15, 42)
(20, 34)
(50, 25)
(100,13)
};
\addlegendentry{LR-Transformer}

\addplot [color=black,mark=star]
coordinates {
(1, 237)
(5,	 184)
(10, 147)
(15, 112)
(20, 98)
(50, 85)
(100,57)
};
\addlegendentry{Transformer}

\end{axis}
\end{tikzpicture}
\caption{The number of uncoordinated slots on the validation set of SNIPS during training.}
\label{fig:un-slots}
\end{figure}

\begin{figure*}[h]
\centering
\begin{tikzpicture}[node distance = 0cm]

\tikzstyle{Word} = [rectangle,minimum width = 1cm, minimum height = 2cm, text centered, draw = white, fill = white,thick]
\tikzstyle{Back} = [rectangle,minimum width = 15cm, minimum height = 7cm, text centered, draw = black, fill = white,thick]

\tikzstyle{every node}=[scale=0.9]
\node(CENTER)[Back]{};
\node(word1)[Word,below of=CENTER,xshift=-6.7cm,yshift=2cm]{play};
\node(word2)[Word,right of=word1,xshift=1cm,yshift=0cm]{the};
\node(word3)[Word,right of=word2,xshift=1.7cm,yshift=0]{video};
\node(word4)[Word,right of=word3,xshift=2.5cm,yshift=0]{game};
\node(word5)[Word,right of=word4,xshift=2.5cm,yshift=0]{the};
\node(word6)[Word,right of=word5,xshift=2.5cm,yshift=0]{genesis};
\node(word7)[Word,right of=word6,xshift=2.5cm,yshift=0]{machine};
\node(TokenSequence)[below of=word4,xshift=1.15cm,yshift=0.4cm]{\textbf{Token Sequence}:};
\node(slot1)[below of=word1,xshift=0cm,yshift=-1.2cm]{O};
\node(slot2)[right of=slot1,xshift=1cm,yshift=0]{O};
\node(slot3)[right of=slot2,xshift=1.7cm,yshift=0]{B-object\_type};
\node(slot4)[right of=slot3,xshift=2.5cm,yshift=0]{I-object\_type};
\node(slot5)[right of=slot4,xshift=2.5cm,yshift=0]{B-object\_name};
\node(slot6)[right of=slot5,xshift=2.5cm,yshift=0]{I-object\_name};
\node(slot7)[right of=slot6,xshift=2.5cm,yshift=0]{I-object\_name};
\node(Answer)[below of=TokenSequence,xshift=0,yshift=-1.2cm]{\textbf{Corret Slots}:};

\node(wslot1)[below of=slot1,xshift=0cm,yshift=-1.2cm]{O};
\node(wslot2)[right of=wslot1,xshift=1cm,yshift=0]{O};
\node(wslot3)[right of=wslot2,xshift=1.7cm,yshift=0]{B-object\_type};
\node(wslot4)[right of=wslot3,xshift=2.5cm,yshift=0]{{\color{red}I-object\_name}};
\node(wslot5)[right of=wslot4,xshift=2.5cm,yshift=0]{{\color{red}B-object\_type}};
\node(wslot6)[right of=wslot5,xshift=2.5cm,yshift=0]{I-object\_name};
\node(wslot7)[right of=wslot6,xshift=2.5cm,yshift=0]{I-object\_name};
\node(Basic)[below of=Answer,xshift=0,yshift=-1.2cm]{\textbf{Basic Model}:};

\node(srslot1)[below of=wslot1,xshift=0cm,yshift=-1.2cm]{O};
\node(srslot2)[right of=srslot1,xshift=1cm,yshift=0]{O};
\node(srslot3)[right of=srslot2,xshift=1.7cm,yshift=0]{B-object\_type};
\node(srslot4)[right of=srslot3,xshift=2.5cm,yshift=0]{I-object\_type};
\node(srslot5)[right of=srslot4,xshift=2.5cm,yshift=0]{{\color{red}B-object\_type}};
\node(srslot6)[right of=srslot5,xshift=2.5cm,yshift=0]{{\color{red}I-object\_type}};
\node(srslot7)[right of=srslot6,xshift=2.5cm,yshift=0]{{\color{red}I-object\_type}};
\node(SR)[below of=Basic,xshift=0,yshift=-1.1cm]{\textbf{SlotRefine}:};

\node(myslot1)[below of=srslot1,xshift=0cm,yshift=-1.2cm]{O};
\node(myslot2)[right of=myslot1,xshift=1cm,yshift=0]{O};
\node(myslot3)[right of=myslot2,xshift=1.7cm,yshift=0]{B-object\_type};
\node(myslot4)[right of=myslot3,xshift=2.5cm,yshift=0]{I-object\_type};
\node(myslot5)[right of=myslot4,xshift=2.5cm,yshift=0]{B-object\_name};
\node(myslot6)[right of=myslot5,xshift=2.5cm,yshift=0]{I-object\_name};
\node(myslot7)[right of=myslot6,xshift=2.5cm,yshift=0]{I-object\_name};
\node()[below of=SR,xshift=0,yshift=-1.2cm]{\textbf{\modelname{}}:};
\end{tikzpicture}
\caption{Case study for the SF task.}
\label{fig:case-study}
\end{figure*}

We first visualize the number decrease of uncoordinated slots in the training process. As shown in Figure \ref{fig:un-slots}, the number of uncoordinated slots drops slow and inefficient for the basic model. 
For our LR-Transformer, the number of uncoordinated slots drops significantly faster. Ten epochs training of our model is already better than 100 epochs training of the basic model on this problem.
Compared with SlotRefine, our model also drops much faster and achieves better convergence after 5 epochs of training. 

\begin{table}[h]
\centering
\resizebox{\columnwidth}{!}{
\begin{tabular}{l|cccc}
\toprule
Model   &slot error   & Unc. error &BI error &IB error\\
\hline
Basic model         & 169    & 57(33.7\%)  & 31    & 26   \\ 
SlotRefine             & 127    & 28(22.0\%)  & 8     & 20    \\ 
LR-Transformer       & 117   & 13(11.1\%)  & 7   & 6  \\
\bottomrule
\end{tabular}
}
\caption{The statistics of slot error on the validation set of SNIPS after training 100 epochs.}
\label{error-analysis}
\end{table}

To further analyze the error in SF tasks in detail, we show the statistics of slot error (i.e., incorrect slots) on the validation set of SNIPS after training 100 epochs. Specifically, we define the errors caused by the uncoordinated slot problem as "Unc. error". The uncoordinated slots includes two cases, i.e., correct `\emph{B-tag}' followed wrong `\emph{I-tag}' and wrong `\emph{B-tag}' following correct `\emph{I-tag}'. We define the first case as "BI error" and the second case as "IB error.", respectively.

The statistics show that the uncoordinated slot problem composes a big part of all slot errors. Without any approach to solving this problem, our basic model encounters 57 uncoordinated slots, composing 33.7\% of all slot errors. In these uncoordinated slots, 31 are caused by the "BI error," while 26 are caused by the "IB error." The proportion of the "BI error" and the "IB error" is almost close.

Compared with the basic model, our LR-Transformer reduces 44 uncoordinated slots. The proportion of uncoordinated slots in all incorrect slot labels drops from 33.7\% to 11.1\%. Moreover, our LR-Transformer efficiently reduces both the BI error and the IB error simultaneously. The reducing proportion between the BI error and the IB error of our model is also close. This is mainly because our SLG effectively obtains sequential dependency while LRM considers both B-tag slots and I-tag slots.  
As a comparison, SlotRefine correct most uncoordinated slots caused by the "BI error" but still suffering "IB error." For SlotRefine, the "IB error" proportion is much higher than "BI error." 

\begin{table*}[t]
\centering
\begin{tabular}{l|ccc|ccc}
\hline
\multirow{2}{*}{\textbf{Model}}  & \multicolumn{3}{c|}{\textbf{ATIS}}    & \multicolumn{3}{c}{\textbf{SNIPS}} \\ 
\cline{2-7} & \multicolumn{1}{c}{Intent} & \multicolumn{1}{c}{Slot} & Overall & \multicolumn{1}{c}{Intent} & \multicolumn{1}{c}{Slot} & Overall \\ 
\hline
Basic model \cite{shaw2018position_relative_attention}   & 96.8      & 94.8      & 85.3      & 96.1      & 92.8      & 82.1 \\
\hline
\modelname{} w/o SLG   & $97.7_{0.047}$      & $95.6_{0.123}$       & $86.1_{0.082}$       & $98.0_{0.094}$       & $93.6_{0.125}$       & $84.9_{0.216}$  \\
\modelname{} w/o LRM*    & $97.0_{0.094}$       & $95.9_{0.169}$       & $86.2_{0.216}$       & $97.6_{0.124}$       & $94.1_{0.249}$       & $86.3_{0.205}$  \\
\modelname{} w/o LRM    & $97.1_{0.047}$      & $95.9_{0.094}$       & $86.3_{0.094}$       & $97.7_{0.047}$      & $94.2_{0.047}$       & $86.5_{0.094}$  \\
\modelname{}* & ${98.1}_{0.047}$  & ${96.1}_{0.163}$  & ${87.1}_{0.141}$  & ${98.7}_{0.081}$  & ${94.6}_{0.169}$  & ${88.2}_{0.169}$  \\
\hline
\modelname{} & $\textbf{98.2}_{0.047}$  & $\textbf{96.1}_{0.047}$ & $\textbf{87.2}_{0.081}$  & $\textbf{98.8}_{0.047}$  & $\textbf{94.8}_{0.094}$  & $\textbf{88.4}_{0.081}$  \\
\hline
\end{tabular}
\caption{Performance (mean and standard deviation of the model repeated 5 times) comparison of each module in our model. Models with * represent not using consistency loss in Eq.\ref{eq_slgloss}, i.e., $\alpha = 0$ in Eq.\ref{eq_slgloss}.}
\label{res_abl}
\end{table*}

We provide an example for the case study. In Figure \ref{fig:case-study}, we notice that the basic model suffers a serious uncoordinated slot problem, including both two cases, i.e., "BI error" and "IB error," respectively.
SlotRefine solves "BI error" but predicts incorrect slot labels in the "IB error" case due to the error propagation from the wrong B-tag label. Our model solves problems of both cases and predicts a correct slot label sequence. 

Above all, our LR-Transformer indeed remedies the problem of the uncoordinated slots, leading to better performance on SF.

\subsection{Ablation Study}
In this section, we do an ablation study to verify the effectiveness of SLG and LRM in detail.

\textbf{Effect of SLG:}
As shown in Table \ref{res_abl}, compared with the basic model, the \modelname{} w/o LRM (i.e., basic model + SLG) enhances the performance with a large margin. Specifically, SLG brings enhancement by 0.3\% for ID, 1.1\% for SF, and 1.0\% for overall on ATIS while 1.6\% for ID, 1.4\% for SF, and 4.4\% for overall on SNIPS. 
We attribute this improvement to the autoregressive structure of SLG, which brings sequential solid dependency information, making prediction more accurate.

Moreover, comparing LR-Transformer with LR-Transformer* and LR-Transformer w/o LRM with LR-Transformer w/o LRM*, we find that utilizing consistency loss function in Eq.\ref{eq_slgloss} brings a slight enhancement. More importantly, the consistency loss effectively reduces the standard deviation, especially on SF and overall. This is mainly because utilizing consistency loss can enhance the prediction consistency between the SLU task and the SLG task, which makes the model more stable. Moreover, compared with SF, the standard deviation on ID tasks is much lower. We consider the reason is ID contains fewer categories and easier to predict. 

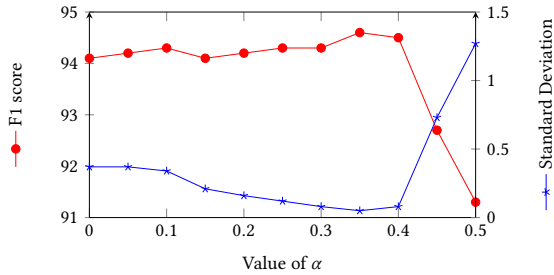
\begin{figure}[h]
\centering
\begin{tikzpicture}[scale = 0.8]
\begin{axis}[
height=5cm,
width=8cm,
xlabel=Value of $\alpha$,
ylabel=\ref{y1} F1 score,
axis y line=left, 
xmin=0,xmax=0.5,
ymin=91.0,ymax=95.0,
ytick pos=left,
xtick = {0,0.1,0.2,0.3,0.4,0.5}
]
\addplot[color=red,mark=*] coordinates {
(0, 94.1)
(0.05, 94.2)
(0.1, 94.3)
(0.15, 94.1)
(0.2, 94.2)
(0.25, 94.3)
(0.3, 94.3)
(0.35, 94.6)
(0.4, 94.5)
(0.45, 92.7)
(0.5, 91.3)
};\label{y1}
\end{axis}

\begin{axis}[
height=5cm,
width=8cm,
axis x line=none,
axis y line=right, 
ylabel=\ref{y2} Standard Deviation,
ymin=0,ymax=1.5,
ytick pos=right,
xmin=0,xmax=0.5,
xtick = {0,0.05,0.1,0.15,0.2,0.25,0.3,0.35,0.4,0.45,0.5}
]
\addplot[color=blue,mark=star] coordinates {
(0, 0.37)
(0.05, 0.37)
(0.1, 0.34)
(0.15, 0.21)
(0.2, 0.16)
(0.25, 0.12)
(0.3, 0.08)
(0.35, 0.05)
(0.4, 0.08)
(0.45, 0.73)
(0.5, 1.27)
};\label{y2}
\end{axis}

\end{tikzpicture}
\caption{Impact of $\alpha$ on the SF performance on the SNIPS validation set.}
\label{fig:alpha}
\end{figure}

We further conduct experiments to study the impact of $\alpha$ on the SF performance. In this experiment, we keep the other settings and change the $\alpha$ from 0 to 0.5. According to the results shown in Figure \ref{fig:alpha}, we find that the F1 scores change little when $\alpha$ increases from 0 to 0.35, but the stand deviation drops continuously. Then, when $\alpha$ is larger than 0.4, both the F1 score and the stand deviation get worse rapidly. We consider the large weight of $\alpha$ makes the wrong prediction in SLU excessively affect the SLG task and bring negative effect since the predicted labels of SLU are not the same as the ground-truth labels.

\begin{figure}[h]
\centering
\begin{tikzpicture}[scale = 0.8]
\begin{axis}[
height=5cm,
width=8cm,
xlabel=Value of $\lambda$,
ylabel=F1 score,
xmin=0,
xmax=1,
ymin=93.0,
ymax=95.0,
ytick pos=left,
xtick={0,0.25,0.5,0.75,1},
]
\addplot[color=red,mark=*] coordinates {
(0, 93.3)
(0.25, 94.2)
(0.5, 94.4)
(0.75, 94.7)
(1.0, 94.1)
};

\end{axis}
\end{tikzpicture}
\caption{Impact of $\lambda$ on the SF performance on the SNIPS validation set.}
\label{fig:lambda}
\end{figure}
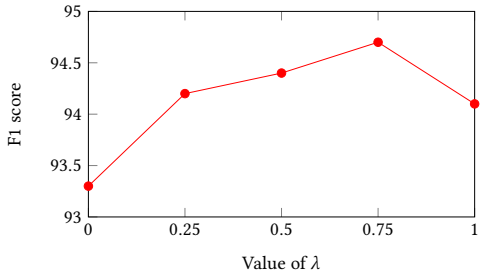

Similar to $\alpha$, we also conduct experiments to study the impact of $\lambda$ on the SF performance. As shown in Figure \ref{fig:lambda}, the F1 score is enhanced notably when $\lambda$ changed from 0 to 0.25, which indicates the effect of our SLG module. Then, the F1 score continues to rise until $\lambda$ reaches 0.75 and begins to drop with the increase of $\lambda$. This observation indicates that we should make a suitable balance between the SLU and SLG  in our final framework.

\begin{table*}[h]
\centering
\begin{tabular}{l|ccc|ccc}
\hline
\multirow{2}{*}{\textbf{Model}}  & \multicolumn{3}{c|}{\textbf{ATIS}}    & \multicolumn{3}{c}{\textbf{SNIPS}} \\ 
\cline{2-7} & \multicolumn{1}{c}{Intent} & \multicolumn{1}{c}{Slot} & Overall & \multicolumn{1}{c}{Intent} & \multicolumn{1}{c}{Slot} & Overall \\ 
\hline
BERT-SLU \cite{chen2019bert}       & 97.5       & 96.1          & 88.2        & 98.6       & 97.0          & 92.8   \\
BERT + Stack-Propagation \cite{qin2019stack}           & 97.5       & 96.1          & 88.6        & 99.0       & 97.0          & 92.9   \\
BERT + SlotRefine  \cite{wu2020slotrefine}  & 97.7       & 96.1          & 88.6        & 99.0       & 97.0          & 92.9   \\ 
\hline
\modelname{}   & 98.2      & 96.1      & 87.2      & 98.4      & 94.8      & 88.4 \\
BERT+SLG & \ \textbf{98.3}$\uparrow$ & \ \textbf{96.2}$\uparrow$ & \ \textbf{88.7}$\uparrow$ & \ \textbf{99.1}$\uparrow$ & \ \textbf{97.1}$\uparrow$ & \ \textbf{93.1}$\uparrow$ \\
\hline
\end{tabular}
\caption{SLU performance of Bert-based models on ATIS and SNIPS datasets. The numbers with $\uparrow$ indicate that the improvement of our model over all baselines is statistically significant with $p < 0.05$ under t-test.}
\label{res_bert}
\end{table*}
\textbf{Effect of LRM:}
By comparing \modelname{} w/o SLG (i.e., basic model + LRM) with the basic model, we find that LRM enhances the performance by 0.9\%(ID), 0.8\%(SF), and 0.8\%(Overall) on ATIS and 1.9\%(ID), 0.8\%(SF), and 2.8\%(Overall) on SNIPS, which shows that utilizing LRM alone still enhances SLU performance on both two tasks. We attribute the improvement to the direct utilization of ID and SF preliminary results. The semantic information from both slot labels and ID labels enhances the performance when predicting.


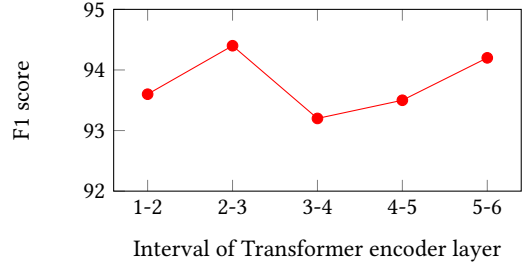
\begin{figure}[h]
\centering
\begin{tikzpicture}[scale = 1]
\begin{axis}[
height=4cm,
width=7cm,
xlabel=Interval of Transformer encoder layer,
ylabel=F1 score,
ymin=92.0,
ymax=95,
ytick pos=left,
symbolic x coords={1-2,2-3,3-4,4-5,5-6},
x tick label style={/pgf/number format/1000 sep={}},
]
\addplot[color=red,mark=*] coordinates {
(1-2,93.6)
(2-3,94.4)
(3-4,93.2)
(4-5,93.5)
(5-6,94.2)
};

\end{axis}
\end{tikzpicture}
\caption{SF performance comparison of using LRM in different Transformer encoder intervals on SNIPS validation set. 1-2 represents the interval of the 1st and the 2nd Transformer encoder layer, and so on.}
\label{fig:LRMpos}
\end{figure}

As we introduced in section \ref{sec:LRM}, LRM can be utilized in each two Transformer encoder layer intervals. Motivated by finding the best place to operate LRM, we conduct the experiment to compare SF performance on the validation set of SNIPS when using LRM in different Transformer encoder intervals. The experiment result is shown in Figure \ref{fig:LRMpos}. 

\begin{figure}[h]
\centering
\begin{tikzpicture}[scale = 1]
\begin{axis}[
height=4cm,
width=7cm,
xlabel=Usage count of LRM,
ylabel=F1 score,
ymin=92.0,
ymax=95.0,
ytick pos=left,
symbolic x coords={1,2,3,4,5},
x tick label style={/pgf/number format/1000 sep={}},
]
\addplot[color=red,mark=*] coordinates {
(1, 94.4)
(2,	94.3)
(3, 93.5)
(4,93.2)
(5,92.1)
};

\end{axis}
\end{tikzpicture}
\caption{SF performance comparison of different LRM usage count on SNIPS validation set.}
\label{fig:LRMnum}
\end{figure}
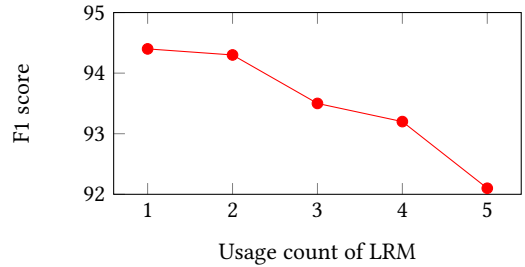

We further conduct experiments to evaluate SF performance for different usage count of LRM. As shown in Figure \ref{fig:LRMnum}, the F1 score on the validation set achieves a satisfying level when employing LRM once or twice and drops significantly with increasing usage count.
This phenomenon indicates that employing LRM one time is enough, and overuse LRM will negatively influence SLU prediction. We have introduced the reason in section 3.3, using LRM once in the whole Transformer structure already considers the interaction between ID and SF. Overusing LRM  will make the Transformer layers lose their own feature and negatively affect the final performance due to the utilization of the SLU classifier in LRM.

\subsection{Effect of Pretraining}
In this section, we conduct experiments to evaluate the ability of our model to combining the pretrained model. 

Recently, some work builds their models based on large-scale pre-trained model BERT \cite{BERT}, which utilized billions of external corpus and tremendous model parameters. Since the number of BERT parameters is much more than ours, it is unfair to compare the performance of our model with them directly. Thus, we also combine our approaches with BERT to highlight the effectiveness of \modelname{}.

To make better use of the pre-trained results of BERT, we retain the weights of the original BERT model and do not change the inner structure of BERT. Therefore, we only combine our SLG module with BERT. In practice, we employ BERT as the encoder and utilize the same Transformer decoder as we introduced in section 3.2. Then we fine-tune the complete model on the SLU dataset. 

As shown in Table \ref{res_bert}, our BERT+SLG outperforms all previous BERT-based models on all evaluation metrics. Compared with BERT-SLU, SLG brings an enhancement of 0.8\%(ID), 0.1\%(SF), and 0.5\%(Overall) on ATIS dataset and 0.5\%(ID), 0.1\%(SF), and 0.3\%(Overall) on SNIPS dataset. All of these improvements of our model are statistically significant with $p<0.05$ under t-test. The results show that our SLG approach is still useful for BERT-based models. This phenomenon also indicates that our SLG module is portable and effective.

\begin{table}[h]
\centering
\resizebox{\columnwidth}{!}{
\begin{tabular}{l|cc|cc}
\hline
\multirow{2}{*}{Model} & \multicolumn{2}{c|}{\textbf{ATIS}}          & \multicolumn{2}{c}{\textbf{SNIPS}}         \\ \cline{2-5} 
                       & \multicolumn{1}{c}{Latency } & Speedup & \multicolumn{1}{c}{Latency } & Speedup \\ 
\hline
BERT+Stack-Propagation      & 220.11ms   & 1.00$\times$  & 225.87ms  & 1.00$\times$   \\
BERT-SLU        & 48.90ms   & 4.50$\times$  & 49.59ms  & 4.55$\times$ \\
BERT+SlotRefine             & 97.81ms    & 2.25$\times$  & 99.19ms   & 2.27$\times$   \\ 
\hline
\modelname{}    & 13.11ms   & -  & 14.01ms  & -  \\
BERT+SLG     & 48.89ms   & 4.50$\times$  & 49.61ms  & 4.55$\times$ \\
\hline
\end{tabular}
}
\caption{Latency of Bert-based SLU models. ``Latency'' is the average inference time without minibatching. ``Speedup'' is compared against the existing SOTA model BERT+Stack-Propagation.}
\label{res_speed2}
\end{table}

We also record the inference time for each SLU model when employing BERT.
As shown in Table \ref{res_speed2}, our BERT+SLG also achieves significant speedup ($\times$4.50 on ATIS; $\times$4.55 on SNIPS) against the autoregressive SOTA model Stack-Propagation. Meanwhile, comparing with the existing non-autoregressive model SlotRefine, our model can reduce 50\% inference time, as our model does not need to run BERT two times. 
It is worth noting that our model consumes almost the same time as BERT-SLU \cite{chen2019bert} because our SLG module works only in the training period and wastes no extra time during inference.

From Table \ref{res_bert}, we further find that the improvement of BERT-based models on the SNIPS dataset is much more obvious than that on the ATIS dataset. Considering that SNIPS contains much more Out-of-Vocabulary (OOV) tokens, we attribute the improvement of BERT to alleviating the OOV problem by the WordPiece encoding \cite{schuster2012japanese}. Besides, although BERT can bring enhancement, the trade-off for this enhancement is 3.77 times of inference latency when comparing BERT+SLG with LR-Transformer in Table \ref{res_speed2}.

To sum up, our designed SLG task is well combined with the pre-trained model BERT, and BERT+SLG outperforms all existing models on both performance and inference latency.

\section{Conclusion}
In this paper, we propose a fast and accurate non-autoregressive model: \modelname{}. To address the sequential dependency information among tokens, we design the SLG task, which effectively enhances the performance of SF through multi-task learning and costs no extra inference time. 
We also design a Layered Refine Mechanism, which guides the final prediction via the interaction of predicted results between Transformer layers. LRM explicitly introduces the correlation between ID and SF into the model with a little cost of time.
Experiments on two public datasets indicate that our model significantly improves performance while substantially accelerate the inference speed. 

\section{Acknowledgements}
This work is supported by Guangdong Key Lab of AI and Multi-modal Data Processing, Chinese National Research Fund (NSFC) Project No. 61872239; BNU-UIC Institute of Artificial Intelligence and Future Networks funded by Beijing Normal University (Zhuhai) and AI-DS Research Hub, BNU-HKBU United International College (UIC), Zhuhai, Guangdong, China. 

\bibliographystyle{ACM-Reference-Format}
\bibliography{cikm2021}

\end{document}